\documentclass[runningheads]{llncs}
\usepackage{graphicx}

\usepackage{tikz}
\usepackage{comment}
\usepackage{amsmath,amssymb} 
\usepackage{color}
\usepackage{booktabs}
\usepackage{algorithm}
\usepackage{appendix}
\usepackage[noend]{algorithmic}

\usepackage[accsupp]{axessibility}  

\begin{document}
\pagestyle{headings}
\mainmatter
\def\ECCVSubNumber{2621}  
\newcommand{\RM}[1]{\textcolor{blue}{#1}}
\title{3D Instances as 1D Kernels\thanks{Y. Wu and M. Shi contributed equally. Z. Cao is the corresponding author.}}

\titlerunning{3D Instances as 1D Kernels}
%

\author{Yizheng Wu\inst{1}
Min Shi\inst{1} 
Shuaiyuan Du\inst{1} 
Hao Lu\inst{1} 
Zhiguo Cao\inst{1} 
Weicai Zhong\inst{2}}

\authorrunning{Wu et al.}
%
\institute{ Key Laboratory of Image Processing and Intelligent Control, Ministry of Education \\ 
School of AIA, Huazhong University of Science and Technology, China \\
 \and
Huawei CBG Consumer Cloud Service Search \& Maps BU\\
\email{\{yzwu21,min\_shi,sydu,hlu,zgcao\}@hust.edu.cn}  $\ $ \email{zhongweicai@huawei.com}}
\maketitle

\begin{abstract}
We introduce a 3D instance representation, termed \textit{instance kernels}, where instances are represented by one-dimensional vectors that encode the semantic, positional, and shape information of 3D instances. We show that instance kernels enable easy mask inference by simply scanning kernels over the entire scenes, avoiding the heavy reliance on proposals or heuristic clustering algorithms in standard 3D instance segmentation pipelines. The idea of instance kernel is inspired by recent success of dynamic convolutions in 2D/3D instance segmentation. However, we find it non-trivial to represent 3D instances due to the disordered and unstructured nature of point cloud data, \textit{e.g.}, poor instance localization can significantly degrade instance representation. To remedy this, we construct a novel 3D instance encoding paradigm. First, potential instance centroids are localized as candidates. Then, a candidate merging scheme is devised to simultaneously aggregate duplicated candidates and collect context around the merged centroids to form the instance kernels. Once instance kernels are available, instance masks can be reconstructed via dynamic convolutions whose weights are conditioned on instance kernels. The whole pipeline is instantiated with a dynamic kernel network (DKNet). Results show that DKNet outperforms the state of the arts on both ScanNetV2 and S3DIS datasets with better instance localization. Code is available: \url{https://github.com/W1zheng/DKNet}. 


\end{abstract}

\section{Introduction}
3D Instance segmentation aims to predict point-level instance labels~\cite{maskrcnn,pointgroup}. Standard approaches heavily rely on proposals~\cite{bonet,3dmpa,gicn} or heuristic clustering algorithms~\cite{pointgroup,hais}. In this work, we show that instance masks can be reconstructed by scanning a scene with \textit{instance kernels}, a representation for 3D instances, which simultaneously 
encodes the positional, semantic, and shape information of 3D instances. 

\begin{figure}[!t]
\centering
\includegraphics[width=1\linewidth]{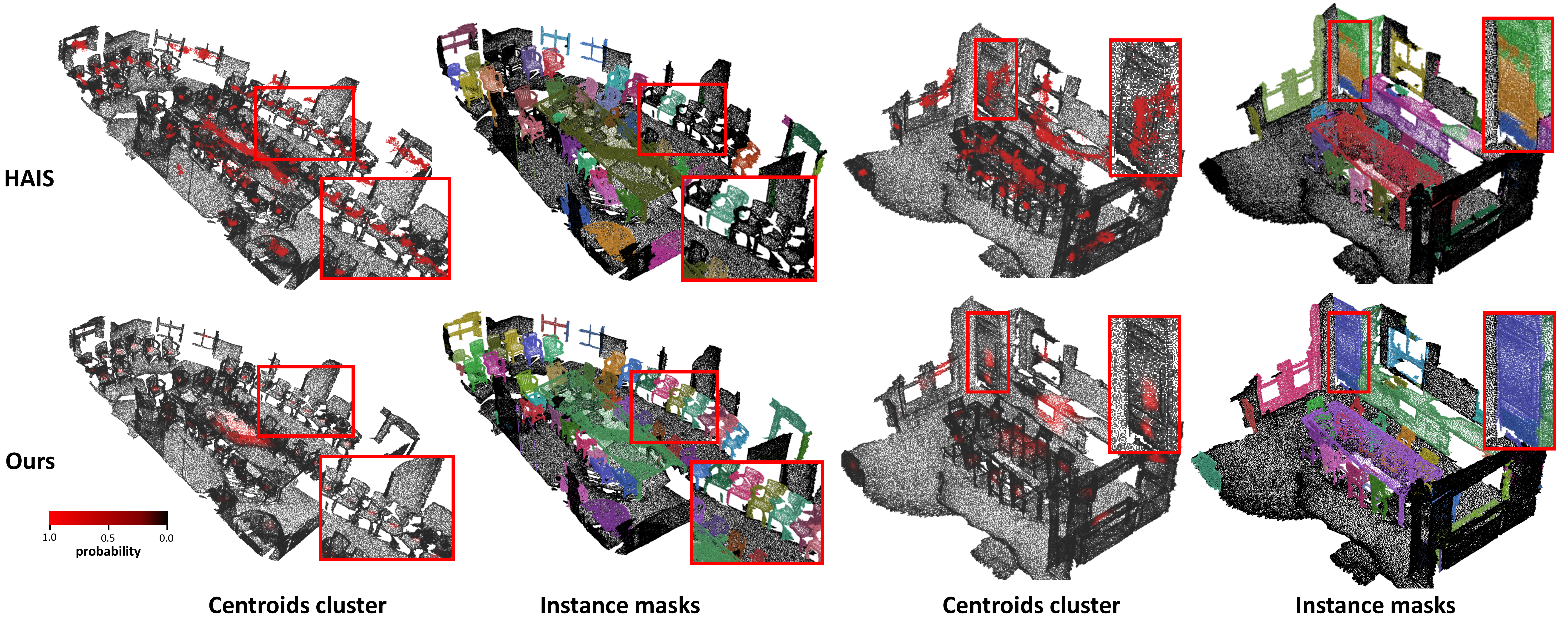}
\caption{\textbf{Comparison of inferred centroid clusters and instance masks.} Compared with HAIS~\cite{hais}, our DKNet generates more focused centroid clusters that can guide precise localization such that small and close instances can be discriminated and large instances have consistent predictions. 
Best viewed by zooming in and in color.
}
\label{fig:fig-paradigm-comparison}
\end{figure}

3D instance representation addresses two fundamental problems: i) how to localize an instance precisely and ii) how to aggregate features effectively to depict the instance. Unlike 2D instances that can be directly encoded via grid sampling~\cite{solov2} or dynamic kernel assigning~\cite{knet}, in the 3D domain, the disordered and unstructured nature of point cloud data renders difficulties for precise instance localization and reliable representation; and top-performing approaches~\cite{hais,sstnet,pointgroup} implicitly localize instances with centroid offsets~\cite{votenet}, which only provides coarse information for instance representation, as shown in Fig.~\ref{fig:fig-paradigm-comparison}.
Our instance kernel also draws inspiration from DyCo3D~\cite{dyco3d}, which first applies dynamic convolution into 3D instance segmentation. However, DyCo3D is built upon the existing bottom-up segmentation pipeline~\cite{pointgroup}, leaving the fundamental problems of instance encoding unsolved.
To alleviate the difficulties above, we design a novel instance encoding paradigm that efficiently localizes different instances and encodes the semantic, positional, and shape information of instances into \textit{instance kernels} for mask generation. 

We further incorporate the kernel encoding paradigm into a dynamic kernel network (DKNet) for 3D instance segmentation.
To localize instances, as shown in Fig.~\ref{fig:fig-paradigm-comparison}, DKNet predicts a centroid map 
for instances and extracts centroids via a customized Non-Maximum Suppression (NMS) operator with local normalization. 
Observing that duplicated candidates may be predicted for a single instance (especially for large ones), we design an iterative aggregation mechanism to merge duplicated candidates guided by a predicted merging score map. The score map indicates the probability whether each paired candidates should be merged. Afterwards, the merged instances are encoded into instance kernels by adaptively fusing the point features around the localized instance centroids. Finally, instance masks can be reconstructed with a few convolution layers, whose weights are conditioned on the generated instance kernels.

We evaluate DKNet on two popular 3D instance segmentation datasets, including ScanNetV2~\cite{scannet} and S3DIS~\cite{s3dis}. The results show that DKNet outperforms previous state-of-the-art approaches, ranking the first AP among published methods on the ScanNetV2 online leaderboard.\footnote[1]{\url{http://kaldir.vc.in.tum.de/scannet\_benchmark/semantic\_instance\_3d}} Thanks to the instance kernels and the specially designed instance localization pipeline, DKNet can better distinguish instances from dense areas than current top-performing approaches, as shown in Fig.~\ref{fig:fig-paradigm-comparison}. A series of ablation studies also demonstrate that the proposed instance localization and aggregation pipeline can greatly enhance the instance representation. 

Our contributions are two-fold:
\begin{itemize}


\item[-] We extend the idea of dynamic convolution into instance kernel, a comprehensive representation for 3D instances in point clouds;

\item[-] We propose a dynamic kernel network for 3D instance segmentation, with a novel instance kernel encoding paradigm;

\end{itemize}

\section{Related Work}
Here we briefly review the 3D instance segmentation approaches and kernel-based instance segmentation. 

\noindent\textbf{Proposal-Based 3D Instance Segmentation.}
Proposal-based approaches~\cite{GSPN} assign instances with proposals and the instance masks are generated upon proposals. 3D-BoNet~\cite{bonet} directly predicts the bounding boxes, within instance masks are generated. 3D-MPA~\cite{3dmpa} samples proposals from predicted centroids; masks of proposals are then clustered to form the instance masks. GICN~\cite{gicn} simultaneously predicts the centroids and sizes of instances to obtain bounding box proposals. Predictions from proposal-based approaches show good objectness, while two major drawbacks exist: 1) the multi-stage training and the proposal generation process introduce large computational overhead; 2) the results highly rely on the proposals. 

\noindent\textbf{Propoal-Free 3D Instance Segmentation.}
Proposal-free approaches cluster points into instances in a bottom-up manner. SSTNet~\cite{sstnet} models the entire scene by constructing a tree of superpoints and uses top-bottom traversal to aggregate nodes and form instance masks. PointGroup~\cite{pointgroup} clusters points using semantic labels and centroid offsets as clues. PE~\cite{pe} encodes points into an embedding space where points from the same instances are close. Then clustering are performed in this embedding space. Considering that clustering points into instances with various sizes in one shot is difficult, HAIS~\cite{hais} proposes a novel hierarchical clustering pipeline to gradually refine the aggregation results. However, even with the implicit guide of object signals, the objectness of predictions is still low, as shown in Fig.~\ref{fig:fig-paradigm-comparison}. Hence, directly adding kernel-based dynamic convolution modules upon existing proposal-free approaches cannot bring out the best of kernel-based instance segmentation paradigm. 

\noindent\textbf{Kernel-Based Instance Segmentation.}
Kernel-based instance segmentation uses instance-aware kernels to scan the whole scene to reconstruct instance masks, the pivot of which is to represent or associate instances with different kernels. After obtaining the kernels, the common solutions are scanning the scene via dot product or dynamic convolution~\cite{dfn,lu2022index}. CondInst~\cite{condinst} predicts instance proposals by object detection and encodes the proposal features into kernels. K-Net~\cite{knet} associates a fixed number of kernels with instances in certain regions and dynamically updates them. SOLOv2~\cite{solov2} partitions the feature maps into grids and generates a kernel for each grid. However, representing instances as kernels in the 3D domain is non-trivial due to the disordered and unstructured nature of point cloud data. 
DyCo3D~\cite{dyco3d} first introduce the kernel-based paradigm in 3D instance segmentation, which is built upon existing bottom-up approach~\cite{pointgroup}. However, they focus on the concrete implementation of dynamic convolutions, and bypass the core of the kernel-based paradigm: how to encode instances into kernels? In this work, we further explore the underlying relation between the discriminative representation of instances and effective kernel-based segmentation, resulting in a novel localize-then-aggregate instance kernel encoding paradigm.

\section{Dynamic Kernel Network for 3D Instance Segmentation}
\subsection{Overview}
\begin{figure}[!t]
\centering
\includegraphics[width=1\linewidth]{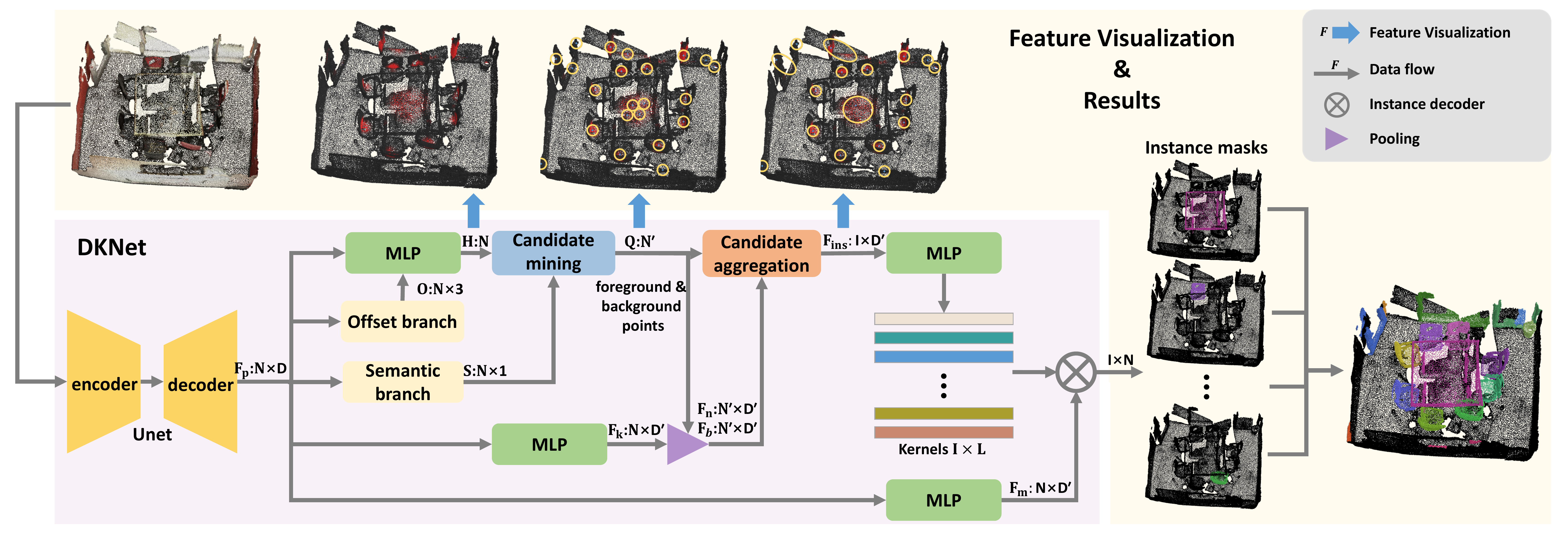}
\caption{\textbf{Pipeline of the Dynamic Kernel Network.} 
}
\label{fig: pipeline}
\end{figure}

As illustrated in Fig.~\ref{fig: pipeline}, at the core of our dynamic kernel network (DKNet) is to encode instances into discriminative instance kernels. The encoding process consists of three key stages: 1) processing raw point clouds with a UNet-like backbone and predicting point features, centroid offsets, and semantic masks; 2) localizing centroids for instances with a candidate mining branch; 3) merging duplicated candidates and collecting context around instance centroids to form instance kernels. 
Once the instance kernels are acquired, the instance masks can be obtained by processing the point cloud features with a few convolution layers, whose weights are conditioned 
on instance kernels.

\subsection{Point-wise Feature Extraction}
Following recent proposal-free approaches~\cite{pointgroup,hais,sstnet}, we adopt the backbone from PointGroup~\cite{pointgroup} for feature extraction. Given the raw point cloud $P\in \mathbb{R}^{N \times 6}$ with $N$ points, a 3D UNet-like~\cite{unet} backbone with sparse convolutions~\cite{ssc} outputs point features $F_p \in \mathbb{R}^{N \times D}$. $F_p$ is then fed to a semantic branch which predicts semantic mask $S \in \mathbb{R}^{N\times C}$, and additionally, a centroid offset branch which infers the offset $O \in \mathbb{R}^{N\times 3}$ of each point to the corresponding instance centroid. 
The semantic branch is a Multi-Layer Perceptron (MLP) with $\tt softmax$ activation at the output layer. Cross-entropy loss and multi-class dice loss~\cite{dice} are used to supervise the training of this branch. Similar to the semantic branch, the centroid offset branch maps $F_p$ into offsets $O$. For each point $P_i$, $O_i \in \mathbb{R}^3$ is a vector pointing to the centroid of instance that covers this point. Further details of the backbone can be referred to the supplementary.   

\subsection{Finding Instances}
\label{sec:candidate-mining-branch}
To generate the kernel for each instance, we should first find all the instances. However, the top-performing proposal-free approaches predict centroid offsets as implicit object signals, which is rather coarse to precisely localize instances, as shown in Fig.\ref{fig:fig-paradigm-comparison}. Hence, learning from proposal-based approaches, we propose a candidate mining branch to generate centroid maps, followed by a searching algorithm to localize the instance candidates.

\begin{figure}[!t]
\centering
\includegraphics[width=1\linewidth]{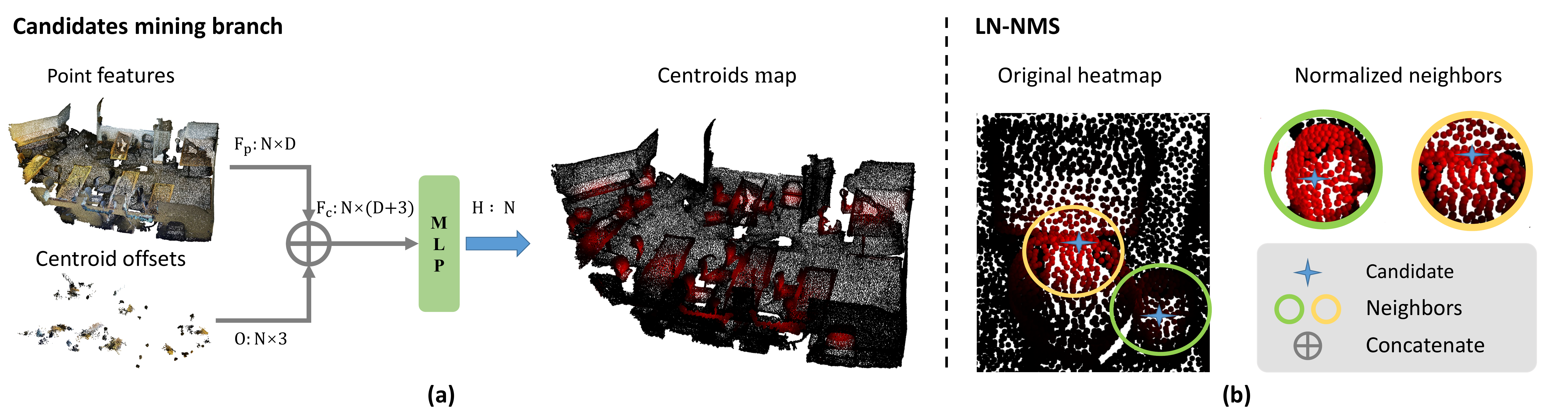}
\caption{\textbf{Centroid mining branch.} (a) The input of candidate mining branch; (b) The customized Non-Maximum Suppression with local normalization. 
}
\label{fig:mining}
\end{figure}

As shown in Fig.~\ref{fig:mining}, point features $F_p \in \mathbb{R}^{N\times D}$ and centroid offsets $O\in \mathbb{R}^{N\times3}$ are jointly concatenated to form the input $F_c \in \mathbb{R}^{N\times(D+3)}$ for the centroid mining branch. Then $F_c$ is fed into an MLP with $\tt softmax$ activation at the output layer to obtain sharp centroid heatmap $H\in \mathbb{R}^{N}$.
Each element $H_i$ indicates the probability of the $i^{th}$ point being an instance centroid.

During training, we place a 3D Gaussian kernel on every instance centroid to form a pseudo ground truth heatmap as \mbox{$\hat{H}_{i} = \mathrm{exp}(-\alpha \cdot {d_i^2}/{r_i^2})\,,$} 
where $d_i$ denotes the distance between point $i$ to the centroid of the instance covering it. $r_i$, which controls the variance of the Gaussian kernel, equals the maximum side length of the axis-aligned bounding box of the corresponding instance. Hence, the Gaussian kernels are geometry adaptive w.r.t.\ the size of different instances. $\alpha$ is set to $25$ to keep the average of $\hat{H}$ around $0.1$. To supervise the training, the loss function $\mathcal{L}_{center}$ for the candidate mining branch is defined as: 
\begin{equation}
    \mathcal{L}_{center} = \frac{1}{\sum_{i=1}^N \mathbb{I}(P_i)}\sum_{i=1}^N\vert H_i - \hat{H}_{i}\vert \cdot \mathbb{I}(P_i)\,,
\end{equation}
where $\mathbb{I}(P_i)$ is an indicator function that outputs 1 when $i^{th}$ point belongs to an instance, otherwise outputs $0$.

With the predicted heatmap $H$, we iteratively search the local maximum as instance candidates with a customized local normalized NMS (LN-NMS) strategy. During each iteration, the algorithm localizes the point with the highest centroid score among the foreground points; the centroid scores of other points in its neighbor with radius $R$ are then normalized via the division of the maximum value in this $R$-radius neighbor. If the normalized centroid score is larger than a threshold $T_\theta$, this point will be considered a candidate and all other points within its $R$-radius neighbor are suppressed and excluded in the next iteration, no matter whether the point is chosen as the candidate. We set $T_\theta=0.5$ and the radius $R=0.3m$ according to the average size of instances in ScanNet~\cite{scannet}. The iteration ends when no point remains or $N_\theta$ candidates have been found. $N_\theta$ is empirically set to $200$. Finally, a candidate 
set $Q \in \mathbb{R}^{N'}$ can be collected, where $N'$ denotes the number of candidates. Refer to supplementary for more details. 

\subsection{Representing Instances as Kernels}
After localizing the instance centroids, we represent these candidates as instance kernels. We expect that one kernel is extracted for one instance, and the kernel should be discriminative. Therefore, we design a duplicated candidate aggregation strategy that simultaneously eliminates extra candidates and adaptively fuses features around candidates for instance representation.
 
\subsubsection{Aggregating Duplicate Candidates.}
We judge whether two candidates should be aggregated based on the context of each candidate. For each raw candidate, we use the features from its ``foreground points'', ``background points'' to describe the context. The ``foreground points'' denote points with the same semantic label within a $R$-radius neighbor of each candidate, while the ``background points'' denote all the points with different semantic labels within a $2R$-radius neighbor of each candidate. To aggregate features from the foreground and background points, we first process point feature $F_p$ with an MLP for dimensionality reduction. Then, the output features of the MLP w.r.t.\ ``foreground points'' and ``background points'' 
are averaged and respectively form the descriptive feature $F_n \in \mathbb{R}^{N'\times D'}$ and the background feature $F_b \in \mathbb{R}^{N'\times D'}$ for each candidate. As the above two features only encode semantic and shape information, we concatenate them with the shifted coordinate (add the raw coordinates with the centroid offset vectors)
of each candidate as positional information to form the aggregation feature $F_a \in \mathbb{R}^{N'\times (2D'+3)}$ for duplicated candidates aggregation.

\begin{figure}[!t]
\centering
\includegraphics[width=1\linewidth]{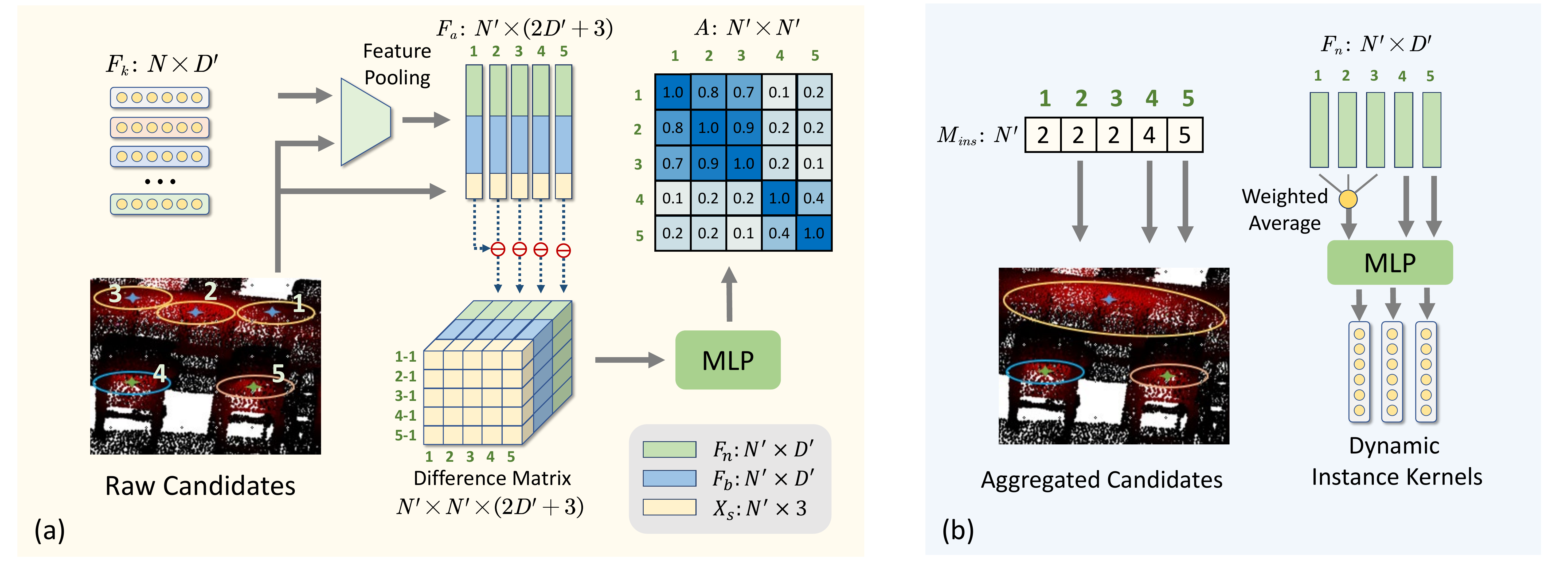}
\caption{\textbf{Candidates aggregation module.} (a) The process of predicting the merging score map; (b) The process of generating instance kernels.}
\label{fig:aggregation}
\end{figure}

As shown in Fig.~\ref{fig:aggregation}, for each candidate, its aggregation feature $F_{a,i}$ is subtracted from the aggregation feature of all other candidates. By repeating this process for every instance, a candidate difference matrix reflecting how similar each pair of candidates will be generated. Taking the absolute values of the difference matrix as inputs, an MLP with $\tt sigmoid$ function outputs the merging score map $A \in \mathbb{R}^{N'\times N'}$, where $A_{ij}$ indicates the probability that the $i^{th}$ and $j^{th}$ candidates shall be merged because they can belong to the same instance. 

Once the merging score map $A$ is obtained, a simple greedy algorithm will be used to iteratively merge candidates. We first initialize an instance centroid map $M_{ins}\in \mathbb{R}^{N}$ where $M_{ins,i}=i$. $M_{ins}$ records the indices of instance centroids that each candidate belongs to, and we define candidates with the same index as an instance \textit{group}. Before aggregation, the instance centroids of candidates are themselves, and each candidate is an instance \textit{group}. During each iteration, if $A_{ij}$ is the maximum in $A$ excluding diagonal elements, all candidates in the $i^{th}$ and $j^{th}$ instance \textit{groups} will be merged. The instance centroid indices of these candidates will also be unified to the index of candidate with the highest centroid score among them. Since candidates within the same instance group can no longer be merged, we then update all merging scores between them to be $0$. The iteration ends when all merging scores are below a predefined threshold, which is set to $0.5$. The candidates with the same index are treated as a predicted instance. After aggregation, supposing $I$ instance groups are generated, the centroid coordinates $C_{ins}\in \mathbb{R}^{I\times 3}$ of new instances are set to the coordinates of the center candidates, while the features of new instances $F_{ins}\in \mathbb{R}^{I \times D'}$ are obtained by a weighted average upon descriptive features of grouped candidates w.r.t.\ their neighbor sizes.
The average can help dynamically aggregate the features of instances with different sizes. As shown in Fig.~\ref{fig:aggregation}, if multiple candidates are predicted on one large instance, information from all the candidates will be propagated to the instance centroids to describe instances.

During training, it is easy to figure out which pairs of candidates are duplicated, and accordingly a ground truth merging map $\hat{A}$ can be generated. The standard binary cross entropy loss (BCELoss) is adopted as the loss function $\mathcal{L}_{aggre}$ for candidate aggregation, which is defined as:
\begin{align}
    \mathcal{L}_{aggre} = BCELoss(A, \hat{A}),
\end{align}
where $\hat{A}_{i,j}=1$ if the candidate $q_i$ and $q_j$ belong to the same ground truth instance, and $0$ otherwise.

Aside from aggregating duplicated candidates, by guiding the network to distinguish whether two candidates need to be merged, the representation ability of point feature $F_p$ can also be enhanced, as the learning of aggregation encourages points from the same instance to be close in the feature space, and vice versa, similar to the idea of contrast learning~\cite{contrast}. 

\subsubsection{Encoding Instance Kernels.} 
After candidate aggregation, all instances in the scene are assigned with centroids, denoted by $C_{ins}$, and corresponding features, denoted by $F_{ins}$. $F_{ins}$ is fed into an MLP to generate instance kernels $\mathcal{W}\in \mathbb{R}^{I\times L}$, where $L$ is the length of instance kernels. In analogous to CondInst~\cite{condinst} and DyCo3D~\cite{dyco3d}, the instance kernels are transformed into the weights for a few convolution layers in the 
instance decoding stage. Hence, $L$ depends on the specific configurations of the 
convolution layers, which can be computed by Eq.~\ref{eq: weight-length}.

\subsection{Generating Masks with Instance Kernels}
The instance kernels, denoted by $\mathcal{W}$, have encoded the positional, semantic, and shape characteristics of instances. To decode instances, the instance kernels are transformed into the weights with a few convolution layers, which are applied to augmented point cloud features to reconstruct instance masks.

To augment the point cloud features, an MLP further extracts the mask feature $F_m \in \mathbb{R}^{N\times D'}$ from the point feature $F_p$. To inject instance-aware positional information into $F_m$, inspired by DyCo3d~\cite{dyco3d}, the offset between each point to the instance centroids are added to $F_m$ before convolution.
\textit{E.g.}, for each point $P_i$, we compute its offset to the centroid of the $k^{th}$ instance as $Z_{k,i}= C_{ins,k} - X_i$. Then, the point decoding feature $F_d\in \mathbb{R}^{N\times (D'+3)}$ for the $k^{th}$ instance is generated by concatenating $F_m$ and $Z_k$ along the channel dimension. Although DyCo3D also generates instance masks via instance-specific kernels, the kernels are only applied to the points within the same semantic category. In contrast, the instance kernels in our approach scan the entire scene, which avoids the reliance of semantic prediction. Hence, the decoding process can correct some errors in semantic prediction. The bottom-up approaches~\cite{pointgroup,hais} cannot correct such errors. The decoding process outputs the instance mask $M\in \mathbb{R}^{I\times N}$ by
\begin{align}
    M_{i} = Conv(F_{d}, \mathcal{W}_i),\quad i\in[1,I]\,,
\end{align}
where $\mathcal{W}_i \in \mathbb{R}^{L}$ is transformed into the weights and biases via two $1\times 1$ convolution layers. The first layer has $16$ output channels with \texttt{ReLU} activation function and the second one has $1$ output channel with $\tt sigmoid$ for mask decoding. To fit the number of parameters of convolution layers, the length of the instance kernel $L$ can be computed by
\begin{align}
\label{eq: weight-length}
    L = (16+3)\times 16(weight)+16(bias) + 16\times 1(weight)+1(bias)=337\,.
\end{align}
The use of dynamic convolution is the same as \cite{dyco3d,condinst}. Implementation details are depicted in the supplementary.

To supervise the generation of instance masks, we first match the predicted instances with actual instances with Hungarian algorithm~\cite{hungarian} according to a cost matrix. Then, we apply the BCELoss and dice loss~\cite{dice} for supervision.
Supposing $M\in \mathbb{R}^{I\times N}$ with $I$ instance masks is generated, and $\hat{M}\in \mathbb{R}^{G\times N}$ with $G$ ground-truth instance masks is provided, the cost matrix $\mathcal{C} \in \mathbb{R}^{I \times G}$ for the Hungarian algorithm is obtained by:
\begin{align}
    \mathcal{C}_{i,j} = \Vert C_{ins,i} - C_{gt,j} \Vert_2 + \mathbb{I}(S_{ins,i}==S_{gt,j})\,,
\end{align}
where $C_{ins}$ and $C_{gt}$ are centroid coordinates of predicted an ground truth instances, respectively, and $S_{ins}$ and $S_{gt}$ are corresponding semantic labels. We determine the instance semantic label $S_{ins}$ by voting within the predicted instance. With the cost matrix, one predicted instance mask is expected to be matched to a instance with the closet centroid and identical semantic label.

After the matching process, the predicted instance mask ${M}$ is assigned with the ground truth instance masks $\hat{M}\in \mathbb{R}^{G\times N}$. Then, BCELoss $\mathcal{L}_{bce}$ and dice loss $\mathcal{L}_{dice}$ are computed by:
\begin{align}
    \label{eq: mask-generation-loss}
    \mathcal{L}_{mask} = \frac{1}{I'}\sum_{k=1}^{I} (BCE(M_k, \hat{M}_k)+ (1- 2\frac{M_k\cdot\hat{M}_k}{\vert M_k\vert+\vert \hat{M}_k\vert}))\cdot \mathbb{I}(iou_k>0.25)\,,
\end{align}
where $\hat{M}_k$ denotes the ground truth instance mask for the $k^{th}$ instance, and $M_k$ denotes the predicted one. $iou_k$ denotes the Intersection-over-Union~(IoU) between $M_k$ and $\hat{M}_k$, and $\mathbb{I}$ is an indicator function. We add the constraints so that the loss will only be computed when instances are correctly matched and $I'$ is defined by $I'=\sum_{k=1}^{I}{\mathbb{I}(iou_k>0.25)}$.

\subsubsection{Inference Post-processing.} During inference, to convert the soft instance masks $M_k$'s into hard instance labels and filter out potentially wrong predictions, a simple yet effective two-stage refinement pipeline is proposed. First, some small duplicated fragments or noise is removed. In the second stage, superpoints~\cite{superpoint} are applied to refine the shapes of generated instance masks. Unlike proposal-free approaches~\cite{pointgroup,dyco3d,hais}, our approach do not need NMS or ScoreNet in post-processing, which is efficient. 

In the first stage, given the predicted soft instance mask $M \in \mathbb{R}^{I\times N}$, we first generate the raw instance label by selecting the label of the instance with the highest score in $M$. Then, we define a coverage score $S_c\in \mathbb{R}^{I}$ by:
\begin{align}
    S_{c,k} = N_{inter,k} / N_{intra,k}\,,
\end{align}
where $N_{inter,k}$ and $N_{intra,k}$ denote the number of ``inter-point'' and ``intra-point'' for the $k^{th}$ instance, respectively. The ``inter-point'' denotes the number of points being assigned to the $k^{th}$ instance in the raw instance label, while the ``intra-point'' denotes the number of points in the soft instance mask of the $k^{th}$ instance that are above a threshold $T_{m,k}$. $T_{m,k}$ is determined by the Otsu algorithm~\cite{otsu} that is adaptive to different instances. The cover score indicates the completeness and independence of each instance prediction. We then multiply the coverage score $S_c$ with $M$ to generate the refined soft instance mask. By taking the instance with the highest score in the refined mask, the final hard instance labels can be obtained. Refer to supplementary for more details on computing $T_{m}$.

In the second stage, we first aggregate the raw point clouds into superpoints~\cite{superpoint}. Points within a certain superpoint should belong to the same instance. Hence, we unify the instance label in each superpoint to be the one that most points belong to. Aside from the instance label $R$, we need to assign a confidence score for each predicted instance that indicates the prediction quality for evaluation. This score is obtained by multiplying the average of instance score and semantic score of the ``intra-point''. 

\setlength{\tabcolsep}{1.5pt}
\begin{table}[!t]
\scriptsize
\begin{center}
\caption{\textbf{Quantitative results on the ScanNetV2 test set.} Refer to supplementary full results with all the $20$ categories.}
\label{table:performance_scan}
\begin{tabular}{@{}l|cc|cccccccccccccc@{}}
\toprule
approaches&\rotatebox{90}{$mAP$}&\rotatebox{90}{$AP@50$}&\rotatebox{90}{bed}&\rotatebox{90}{booksh.}&\rotatebox{90}{cabinet}&\rotatebox{90}{chair}&\rotatebox{90}{curtain}&\rotatebox{90}{desk}&\rotatebox{90}{door}&\rotatebox{90}{otherfu.}&\rotatebox{90}{picture}&\rotatebox{90}{refrige.}&\rotatebox{90}{sofa}&\rotatebox{90}{table}&\rotatebox{90}{window}\\
\midrule
3D-BoNet\cite{bonet}&25.3&48.8&67.2&59.0&30.1&48.4&62.0&30.6&34.1&25.9&12.5&43.4&49.9&51.3&43.9\\
3D-SIS\cite{3dsis}&16.1&38.2&43.2&24.5&19.0&57.7&26.3&3.3&32.0&24.0&7.5&42.2&69.9&27.1&23.5\\
MTML\cite{mtml}&28.2&54.9&80.7&58.8&32.7&64.7&81.5&18.0&41.8&36.4&18.2&44.5&68.8&57.1&39.6\\
3D-MPA\cite{3dmpa}&35.5&61.1&83.3&76.5&52.6&75.6&58.8&47.0&43.8&43.2&35.8&65.0&76.5&55.7&43.0\\
PointGroup\cite{pointgroup}&40.7&63.6&76.5&62.4&50.5&79.7&69.6&38.4&44.1&55.9&47.6&59.6&75.6&55.6&51.3\\
GICN\cite{gicn}&34.1&63.8&\bf{89.5}&80.0&48.0&67.6&73.7&35.4&44.7&40.0&36.5&70.0&83.6&59.9&47.3\\
DyCo3D\cite{dyco3d}&39.5&64.1&84.1&\bf{89.3}&53.1&80.2&58.8&44.8&43.8&53.7&43.0&55.0&76.4&65.7&56.8\\
Occuseg\cite{occuseg}&48.6&67.2&75.8&68.2&57.6&84.2&50.4&52.4&56.7&58.5&45.1&55.7&79.7&56.3&46.7\\
PE\cite{pe}&39.6&64.5&77.3&79.8&53.8&78.6&79.9&35.0&43.5&54.7&54.5&64.6&76.1&55.6&50.1\\
SSTNet\cite{sstnet}&50.6&69.8&69.7&88.8&55.6&80.3&62.6&41.7&55.6&58.5&70.2&60.0&72.0&69.2&50.9\\
HAIS\cite{hais}&45.7&69.9&84.9&82.0&67.5&80.8&75.7&46.5&51.7&59.6&55.9&60.0&76.7&67.6&56.0\\
SoftGroup\cite{softgroup}&50.4&\bf{76.1}&80.8&84.5&\bf{71.6}&86.2&\bf{82.4}&\bf{65.5}&62.0&\bf{73.4}&69.9&\bf{79.1}&\bf{84.4}&\bf{76.9}&59.4\\
\bf{Ours}  &\bf{53.2}&71.8&81.4&78.2&61.9&\bf{87.2}&75.1&56.9&\bf{67.7}&58.5&\bf{72.4}&63.3&81.9&73.6&\bf{61.7}\\
\bottomrule
\end{tabular}
\end{center}
\end{table}
\setlength{\tabcolsep}{1.4pt}
\section{Experiments}
In this section, we first compare the proposed Dynamic Kernel Network (DKNet) with other state-of-the-art approaches on two 3D instance segmentation benchmarks: ScanNetV2\cite{scannet} and S3DIS~\cite{s3dis}. Then, we verify the effectiveness of different components in DKNet via a  controlled ablation study. 

\subsection{Implementation Details}

\noindent \textbf{Training Details.} For data preparation, the coordinates and colors are concatenated together to form $6$D vectors for each point. The network is trained on a single RTX 3090 GPU with a batch size of $4$ for $400$ epochs. We use the AdamW~\cite{adamw} optimizer with an initial learning rate of $0.001$, which is adjusted by a cosine scheduler~\cite{cosine} during training. Weight decay is set to $1e$-$5$. Following previous methods~\cite{pointgroup}, we voxelize the point clouds with the size of $0.02m$ for ScanNetV2 and $0.05m$ for S3DIS. 

The overall training loss combines the loss from the semantic prediction, offset prediction, candidate mining, candidate aggregation branch, and the mask generation process, which can be defined as:
\begin{align}
    \mathcal{L} =\mathcal{L}_{sem}+\mathcal{L}_{off}+\mathcal{L}_{center}+\mathcal{L}_{aggre}+\mathcal{L}_{mask}\,,
\end{align}
where $\mathcal{L}_{sem}$ and $\mathcal{L}_{off}$ are the losses for semantic segmentation and centroid offsets prediction, respectively.

\noindent \textbf{Datasets.} We use ScanNetV2 and S3DIS for training and evaluation. ScanNetV2 includes $1,613$ scenes with $20$ different semantic categories. $1,201$, $312$ and $100$ scenes are selected as the training, validation and test set, respectively. Note that the labels for test set are hidden for a fair comparison. Following official evaluation protocol, we use mean average precisions (mAPs) under different IoU thresholds as the evaluation metrics. $AP@25$ and $AP@50$ denote the average precision scores with IoU thresholds set to $25\%$ and $50\%$. $mAP$ denotes the average of all the $APs$ with IoU thresholds ranging from $50\%$ to $95\%$ with a step size of $5\%$.
\setlength{\tabcolsep}{3pt}
\begin{table}[t]
\centering
\begin{minipage}{0.38\linewidth}
\centering
\caption{\textbf{Object detection results on ScanNetV2 validation set.} We report per-class mAP with an IoU of $25 \%$ and $50 \%$. The IoU is computed on bounding boxes. We evaluate the performance of HAIS with the provided model. ``Ours$^-$" denotes the model without candidates aggregation part.}
\label{table:performance_object}

\begin{tabular}{@{}l|cc@{}}
\toprule\noalign{\smallskip}
approach& $AP@25$& $AP@50$\\
\midrule
VoteNet~\cite{votenet}&58.6 & 33.5\\
3DSIS~\cite{3dsis}& 40.2 & 22.5\\
3D-MPA~\cite{3dmpa}&64.2&49.2\\
DyCo3D~\cite{dyco3d}& 58.9& 45.3\\
HAIS~\cite{hais}&66.0&54.2\\
Ours$^-$&65.4&57.9\\
\bf Ours &\bf 67.4&\bf 59.0\\
\bottomrule
\end{tabular}
\end{minipage}~
\begin{minipage}{0.58\linewidth}
\centering
\caption{\textbf{Quantitative results on S3DIS dataset.} We report mCov, mWCov, mPre, and mRec. approaches with $\ddagger$ marks are evaluated on scenes in Area-5. The others are evaluated via 6-fold cross validation.}
\label{table:performance_s3dis}
\begin{tabular}{@{}l|cccc@{}}
\toprule
approach& mCov& mWCov& mPre& mRec\\
\midrule
PointGroup$^\ddagger$~\cite{pointgroup}& - & - & 61.9 & 62.1\\
DyCo3D$^\ddagger$~\cite{dyco3d}& 63.5& 64.6& 64.3& 64.2\\
SSTNet$^\ddagger$~\cite{sstnet}& -  & -  &65.5&64.2\\
HAIS$^\ddagger$~\cite{hais}&64.3&\bf66.0&\bf71.1&65.0\\
\bf Ours$^\ddagger$& \bf64.7&65.6&70.8&\bf65.3\\\toprule
OccuSeg~\cite{occuseg}& -& - &72.8& 60.3\\
GICN~\cite{gicn}& -& -& 68.5& 50.8\\
PointGroup~\cite{pointgroup}& -& -& 69.6& 69.2\\
SSTNet~\cite{sstnet}& -  & -  &73.5&\bf73.4\\
HAIS~\cite{hais}&67.0&70.4&73.2&69.4\\
SoftGroup~\cite{softgroup}&69.3&71.7&\bf75.3&69.8\\
\bf Ours & \bf70.3&\bf72.8&\bf75.3&71.1\\
\bottomrule
\end{tabular}
\end{minipage}

\end{table}
S3DIS dataset consists of $271$ scenes collected from $6$ different areas with $13$ different object categories. Following previous approaches~\cite{pointgroup,dyco3d,hais,sstnet}, we train and evaluate our approach in two ways: 1) scenes from area-5 are used for testing while scenes in other areas are used for training; 2) 6-fold cross validation where each area is used in turn for testing. 
On S3DIS, with the threshold IoU set to 0.5, we report coverage (mCov), weighted coverage (mWCov), mean precision (mPrec), and mean recall (mRec) as evaluation metrics.

\begin{figure}[!t]
\centering
\setlength{\abovecaptionskip}{0.cm}
\includegraphics[width=1\linewidth]{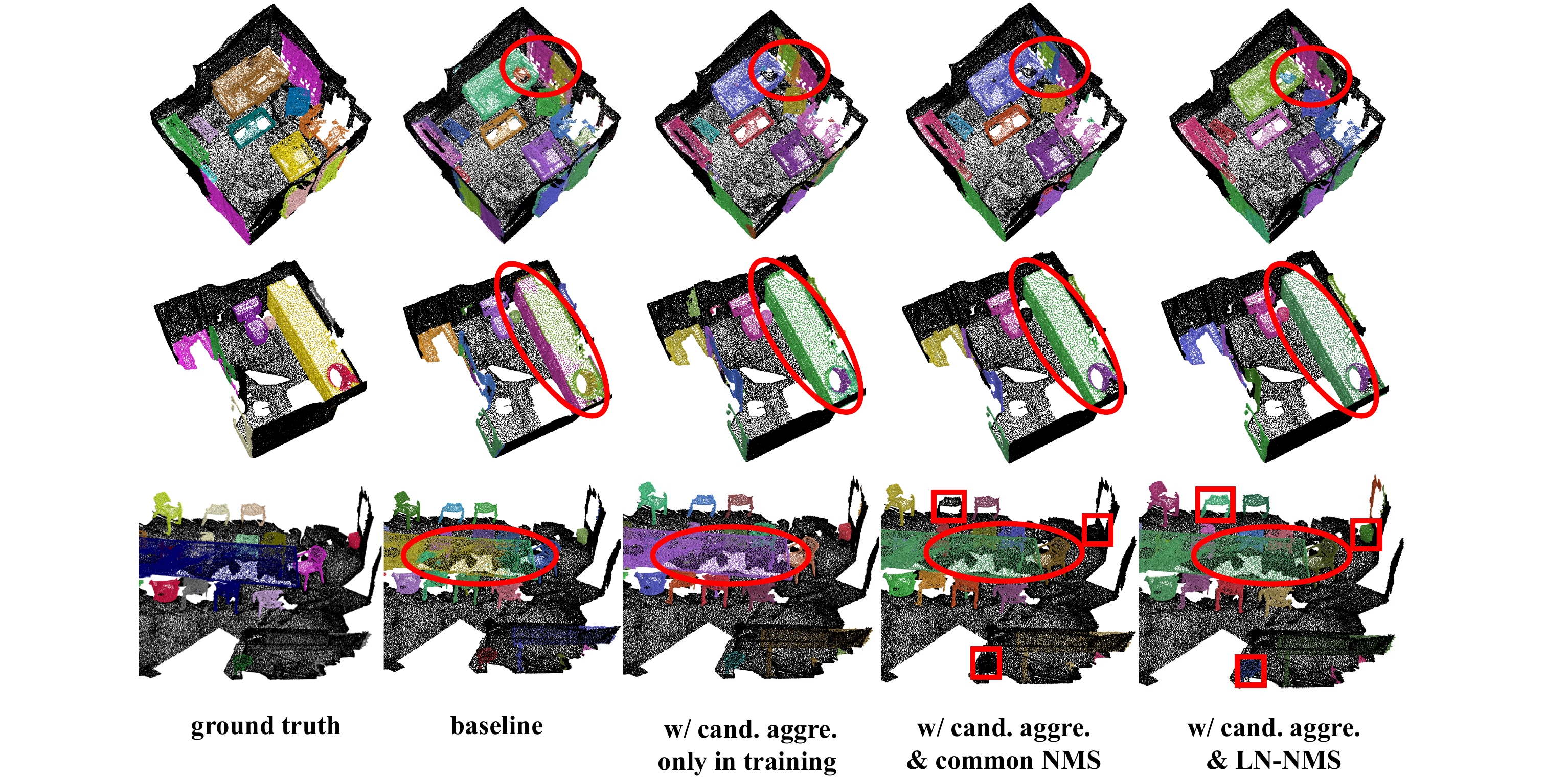}
\caption{\textbf{Qualitative results of our approaches on ScanNetV2 validation set.} We highlight the key details with red marks. Best viewed by zooming in and in color.}
\label{fig:experiments}
\end{figure}
\subsection{Comparison with the state of the arts.}

\noindent \textbf{ScanNetV2.} 
Comparisons with the state of the arts on the ScanNetV2 test set are shown in Table~\ref{table:performance_scan}. 
Our approach achieves an $mAP$ of $53.2\%$, outperforming previous state-of-the-art approaches. The proposed DKNet obtains significant improvement on small instances like chairs or pictures, and competitive results on large instances like beds or tables. We also notice that, compared with a recent well-designed bottom-up method SoftGroup~\cite{softgroup}, DKNet shows inferior $AP@50$.
The plausible reason are two fold. First, we find the DKNet retains relative high AP under strict IoU thresholds ($>$0.6), indicating that the predicted masks can well preserve the instance shapes. However, under lower IoU thresholds ($<$0.6), DKNet becomes less advantageous. Second, our pure data-driven candidate merging process shows mistakes on some difficult scenes, such as bookshelves with vague boundaries, which can be better tackled by well-designed bottom-up clustering. These results suggest that DKNet can be improved with careful design to further boost the potential of the proposed kernel-based paradigm. 

To evaluate the instance localization performance of 3D instance segmentation approaches, we compare different 3D instance approaches under object detection metrics on ScanNetV2 validation set in Table~\ref{table:performance_object}. Predicted masks are converted into axis-aligned bounding boxes following DyCo3D~\cite{dyco3d}. Our approach achieves the best performance in $AP@25$ of $67.4\%$ and $AP@50$ of $59.0\%$, which demonstrate that 
kernels are extracted from solid instance localization results.

\noindent \textbf{S3DIS.} As shown in Table~\ref{table:performance_s3dis}, DKNet is comparable on scenes in area-5, while outperforms other approaches in 3 out of 4 metrics on 6-fold cross validation. Since most categories in S3DIS are large ones like ceiling, wall, and bookcase, the results can reflect the robustness of the proposed instance encoding paradigm on large instances. 

\subsection{Qualitative Evaluation}
We visualize the predicted masks in Fig.~\ref{fig:experiments}. As is shown in column 2, excluding candidate aggregation (baseline) leads to severe over-segmentation as multiple kernels will be generated for one instance. The over-segmentation can be effectively alleviated with the candidate aggregation, which will simultaneously guides the generation of instance-aware features.
Comparing with the common NMS in GICN~\cite{gicn}, the proposed LN-NMS can better localize instances with different sizes, while common NMS omits some small instances (garbage bins in row 2 and chairs in row 3). 
\begin{table}[t]
\centering
\begin{minipage}{0.48\linewidth}
\centering

\caption{Comparison of different candidate mining algorithms on ScanNetV2 validation set.}
\begin{tabular}{@{}l|ccc@{}}
\toprule
Strategy & $AP$ & $AP@50$ & $AP@25 $\\
\midrule
Random & $48.0$ & $63.7$  & $73.5$ \\
NMS & $49.7$ & $64.6$  & $75.7$ \\
LN-NMS & $50.8$ & $66.7$  & $76.9$ \\
\bottomrule
\end{tabular}

\label{table: ablation-candidate-mining}
\end{minipage}~
\begin{minipage}{0.48\linewidth}
\centering
\caption{Comparison of different instance aggregation strategies on ScanNetV2 validation set.}
\begin{tabular}{@{}l|ccc@{}}
\toprule
Phase & $AP$ & $AP@50$ & $AP@25 $\\
\midrule
W/o & $47.7$ & $62.6$  & $74.9$ \\
In training & $48.4$ & $64.5$  & $76.1$ \\
All phases & $50.8$ & $66.7$  & $76.9$ \\
\bottomrule
\end{tabular}

\label{table: ablation-instance-aggregation}
\end{minipage}

\end{table}
\subsection{Ablation Study}
Here, we first compare different candidate mining and aggregation strategies. Then, we analyze how to better represent instances as kernels.

\noindent \textbf{Candidate Mining.}
Here we verify the effectiveness of the proposed LN-NMS algorithm for candidate mining. As in Table~\ref{table: ablation-candidate-mining}, three different ways are compared. \textbf{Random} (row 1) denotes points above a threshold is randomly selected (at most 200) as instance candidates. 
\textbf{NMS} denotes the searching algorithm with common NMS used in GICN~\cite{gicn}.
And the proposed \textbf{LN-NMS} algorithm shown in Sec.~\ref{sec:candidate-mining-branch}. Results in row 1 show that: 1) poor instance localization can significantly degrade instance segmentation performance; 2) even selecting points randomly in the heatmaps as candidates can yield competitive results, which demonstrates the robustness of the subsequent candidate aggregation module. Comparing results in row 2 with row 3, LN-NMS improves $AP@50$ by $2.1\%$.

\noindent \textbf{Instance Candidates Aggregations.} 
Comparison of different candidate aggregation strategies are shown in Table~\ref{table: ablation-instance-aggregation}. ``W/o'' means no candidate aggregation are preformed. ``In training'' means only optimizing aggregation loss during training, while the raw candidates will not be aggregated for inference. ``All phases'' denotes our full approach. When adding the aggregation loss only for supervision (row 2), $AP@50$ increases by $1.9\%$; Motivated by the aggregation loss, there is an instance-clustering trend for the identical instances in the feature space, which is similar to contrast learning~\cite{contrast}. By aggregating the duplicated candidates, the full approach promotes the $AP$, $AP@50$ and $AP@25$ by $3.1\%$, $4.1\%$ and $2.0\%$ comparing with baseline, which demonstrates the effectiveness of candidate aggregation.  

\noindent \textbf{Generating Instance Kernels.} With the raw instance candidates and merging map $M_{ins}$ marking which candidates shall be aggregated, naturally, there are 2 different ways to represent each instance: 1) only using the feature from the candidate with the highest centroids scores; 2) aggregating features from all the merged candidates. As in Table~\ref{table: ablation-instance-representation}, the latter way (default) obtains better results, which demonstrates that aggregating features from different candidates benefits the representation. In addition, we test an ideal representation in row 3 where features are collected from all the points within each instance.
It can be observed that the performances of our instance representation approach are comparable with this ideal representation. This indicates that representative and discriminative instance contexts are obtained in our kernel generation process.

\setlength{\tabcolsep}{4pt}
\begin{table}[t]
\begin{center}
\caption{Comparisons of different candidates aggregation approaches on ScanNetV2 validation set.}
\vspace{-5pt}
\label{table: ablation-instance-representation}
\begin{tabular}{@{}l|ccc@{}}
\toprule
Strategy & $AP$ & $AP@50$ & $AP@25 $\\
\midrule
Candidates elimination & $50.4$ & $65.2$  & $75.8$ \\
Candidates aggregation & $50.8$ & $66.7$  & $76.9$ \\
Full instances & $51.5$ & $67.0$  & $77.0$ \\
\bottomrule
\end{tabular}
\end{center}
\end{table}
\setlength{\tabcolsep}{1.4pt}

\section{Conclusion}
We introduce a 3D instance representation, termed \textit{instance kernels}, which encodes the positional, semantic, and shape information of instances into a 1D vector. We find that the difficulty in representing 3D instances lies in precisely localizing instances and collecting discriminative features. Accordingly, we design a novel 
instance encoding paradigm that first mines centroids candidates for localization. Then, an aggregation process simultaneously eliminates duplicated candidates and gathers features around each instance for representation. We incorporate the instance kernel into a Dynamic Kernel Network (DKNet), which outperforms previous state-of-the-art approaches on public benchmarks. 

\section*{Acknowledgment}
This work was supported in part by the National Key R\&D Program of China (No.2018YFB1305504) and the DigiX Joint Innovation Center of Huawei-HUST.

%
%
\bibliographystyle{splncs04}
\bibliography{egbib}

\appendix   

\setcounter{table}{0}   
\setcounter{figure}{0}

\renewcommand\thesection{A\arabic{section}}
\renewcommand\thesubsection{A\arabic{subsection}}
\renewcommand\thefigure{A\arabic{figure}}
\renewcommand\theequation{A\arabic{equation}}
\renewcommand\thetable{A\arabic{table}}

\section*{Appendix}
\subsection{Backbone and Loss Function}
\begin{figure}[!t]
\centering
\setlength{\abovecaptionskip}{0.cm}
\includegraphics[width=1\linewidth]{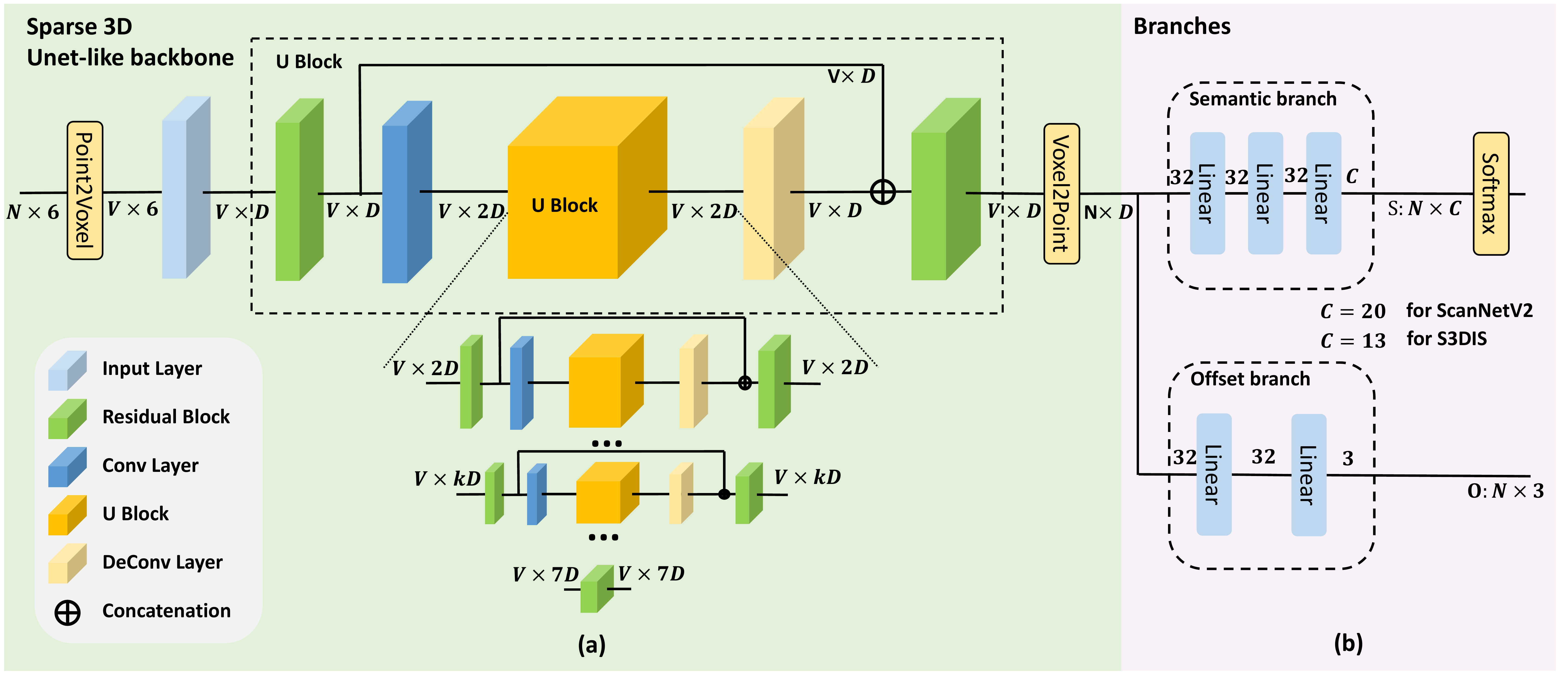}
\caption{\textbf{Detailed network architecture.} (a) Architecture of backbone. (b) Architecture of semantic branch and offset branch. $V$ denotes the number of voxels. The number near linear layer denotes the number of output channels.} 
\label{fig:backbone}
\end{figure}

As mentioned in Sec.~3.2, a 3D UNet-like backbone~\cite{unet} is adopted to extract point features $F_p \in \mathbb{R}^{N \times D}$. And two multi-layer perceptrons (MLPs) are used to predict semantic masks and centroid offsets. We specify the detailed architectures of these three components in Fig.~\ref{fig:backbone}. These three components are adopted from PointGroup~\cite{pointgroup}, which is common for recent top-performing methods~\cite{hais,sstnet}.

\noindent \textbf{Loss for Semantic Branch.} The semantic prediction branch outputs $S \in \mathbb{R}^{N\times C}$, where $C$ is the number of categories. For point $P_i$, $S_i$ denotes the probability of this point belonging to different semantic categories. Given the one-hot ground truth semantic label $\hat{S}_i$, the semantic loss $\mathcal{L}_{sem}$ can be computed as:
\begin{align}
    \mathcal{L}_{sem} = \frac{1}{N} \sum_{i=1}^{N} {CE(S_i, \hat{S}_i) + 1 - \frac{2\sum^{N}_{i=1}{S_i^T\hat{S}_i}}{\sum^{N}_{i=1}{S_i^TS_i}+\sum^{N}_{i=1}{\hat{S}_i^T\hat{S}_i}}, } 
    \label{equa:semantic}
\end{align}
where $CE(x,y)$ denotes the cross entropy loss. The second term in Eq.~\ref{equa:semantic} is the multi-class dice loss~\cite{dice}, which can help address the imbalance between different semantic categories. 

\noindent \textbf{Loss for Offset Branch.} 
The offset branch estimates the centroid offsets for all points, \textit{i.e.}, $O\in \mathbb{R}^{N\times 3}$. Given a point $P_i$, we define the centroid of the instance that covers this point as $C_{p,i}$. Both the Euclidean norm and the direction are considered to measure the difference between the estimated centroid offset vector $O_i$ and the ground truth offset $C_{p,i}-X_i$, where $X_i$ denotes the 3D coordinate of the point $P_i$. Then, the offset loss $\mathcal{L}_{off}$ is computed as:
\begin{align}
    \mathcal{L}_{off} = \frac{1}{N'} \sum_{i=1}^{N} {(\Vert O_i - (C_{p,i} - X_i)\Vert + \frac{O_i \cdot {(C_{p,i} - X_i)}}{\Vert O_i \Vert \cdot \Vert {C_{p,i} - X_i} \Vert}) \cdot \mathbb{I}(P_i), } 
    \label{equa:offset}
\end{align}
where $\mathbb{I}(P_i)$ is an indicator function that outputs $1$ when point $P_i$ belongs to one instance, otherwise outputs $0$. $N'$ denotes the number of points (excluding background points), which can be obtained via $N' = \sum_{i=1}^{N} \mathbb{I}(P_i)$.

\subsection{Algorithms}
\noindent \textbf{Candidate Mining Algorithm}
As mentioned in Sec.~3.3 (line $219$), we design a customized non-maximum suppression algorithm with local normalization (LN-NMS) to localize instance centroids from the predicted heatmaps. The detailed candidate mining process is described in Algorithm~\ref{alg:Searching}. The semantic label $B\in \mathbb{R}^{N}$ denotes the hard semantic label derived from the soft semantic masks $S\in \mathbb{R}^{N\times C}$. $B_i$ equals the category label with the maximum score in $S_i$. Note that, apart from localizing instance centroids, the candidate mining algorithm fetches the ``foreground points'' and ``background points'' of each candidate to describe the candidates for further processing, as mentioned in Sec.3.4 (line $244$). 

\begin{algorithm}[htb]  
  \caption{Candidate mining algorithm $\&$ thresholds $T_\theta$, $Q_\theta$, $R$} 
  \label{alg:Searching}  
  \begin{algorithmic}[1]  
    \REQUIRE  centroids map $H$, coordinates $X$, semantic labels $B$, \\dimension reduced features $F_k$.
    \ENSURE candidates $Q=\{Q_1, Q_2,..., Q_{N'}\}$\\
    neighbor features $F_n = \{F_{n,1}, F_{n,2},..., F_{n,N'}\}$\\
    background features $F_b = \{F_{b,1}, F_{b,2},..., F_{b,N'}\}$
    \STATE initialize an empty candidates set $Q$
    \STATE initialize an empty neighbor features set $F_n$
    \STATE initialize an empty background features set $F_b$
    \STATE initialize an array $f$(available) of length $N$ with all ones
    \STATE initialize counter $k=0$
    \WHILE{$k<T_\theta$ and $f.sum() >0$}
        \STATE initialize distance array $d$ of length $N$ with all zeros
        \STATE initialize neighbors feature $f_n$ of length $C$ with all zeros, counter $k_n=0$
        \STATE initialize background feature $f_b$ of length $C$ with all zeros, counter $k_b=0$
        \STATE /* Candidates mining */
        \STATE set $q=H.argmax()$
        \STATE set $d = \Vert X -X_q\Vert_2$, $f[d<R]=0$
        \IF{$H_q / H[d<R].max() < Q_\theta$}
            \STATE continue 
        \ENDIF
        
        \STATE /* Candidates describing */
        \FOR{$j \in [1,N]$ with $d_j<R$ and $B_j == B_q$}
            \STATE  $f_{n,k}+=f_{d,j}$, $k_n++$
        \ENDFOR
        \FOR{$j \in [1,N]$ with $d_j<2R$ and $B_j != B_q$}
            \STATE  $f_{b,k}+=f_{d,j}$, $k_b++$
        \ENDFOR
        
        \STATE $Q.append(q)$
        \STATE $F_n.append(f_n/k_n)$
        \STATE $F_b.append(f_b/k_b)$
        \STATE $k++$
    \ENDWHILE
    \STATE return $Q, F_n, F_b$
  \end{algorithmic}  
\end{algorithm}

\noindent \textbf{Candidate Merging Algorithm}
As mentioned in Sec.~3.4, to aggregate duplicated candidates with the predicted merging score map $A$
, we design a candidates merging algorithm. Algorithm~\ref{alg:Merging} illustrates this merging process in details.

\begin{algorithm}[htb]  
  \caption{Candidate merging algorithm}  
  \label{alg:Merging}  
  \begin{algorithmic}[1]  
    \REQUIRE merging score map $A$, centroids map $H$, candidates $Q$.
    \ENSURE instance centroid map $M_{ins}\in \mathbb{R}^{N'}$
    \STATE initialize an array $M_{ins} = arange(N')$
    \WHILE{$A.max()>0.5$}
        \STATE $i,j=$ col, row of $A.argmax()$ 
        \STATE $G_i=(M_{ins} == M_{ins}[i])$, $G_j=(M_{ins} == M_{ins}[j])$, $G =$ where$(G_i \cup G_j)$
        \STATE $c=H[Q[G]].argmax()$
        \STATE update $M_{ins}[G] = M_{ins}[c]$
        \FOR{m in G}
            \FOR{n in G}
                \STATE update A[m,n] = 0
            \ENDFOR
        \ENDFOR
    \ENDWHILE
    \STATE return $M_{ins}$
  \end{algorithmic}  
\end{algorithm}

\begin{figure}[!t]
\centering
\setlength{\abovecaptionskip}{0.cm}
\includegraphics[width=1\linewidth]{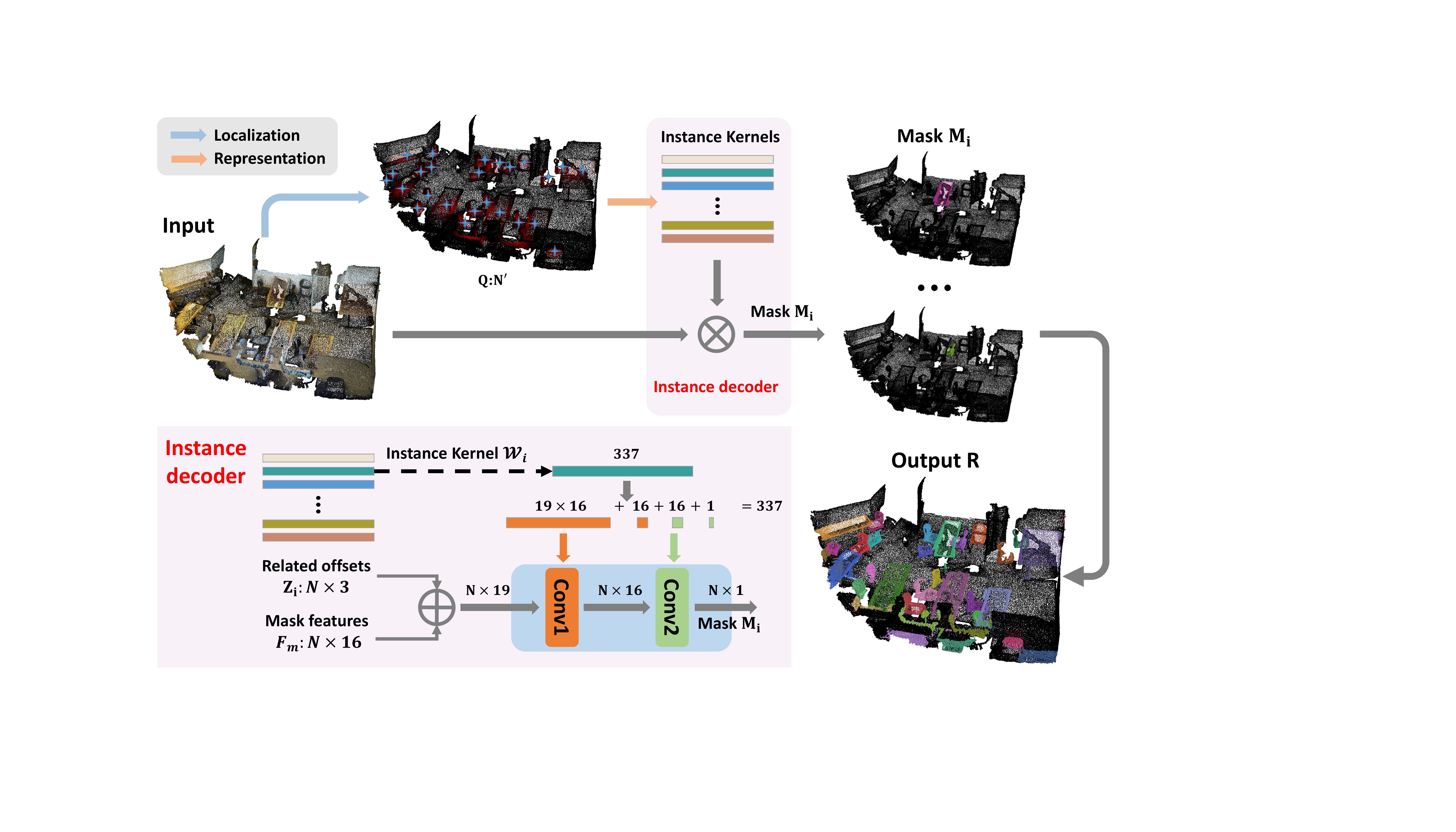} 
\caption{\textbf{Details of instance decoder.} The instance decoder consists of two convolution layers. The elements in instance kernels are sequentially inserted into the weights and biases for convolution layers.}
\label{fig:convolution}
\end{figure}

\subsection{Instance kernels}
\begin{figure}[t]
\centering
\includegraphics[width=0.8\linewidth]{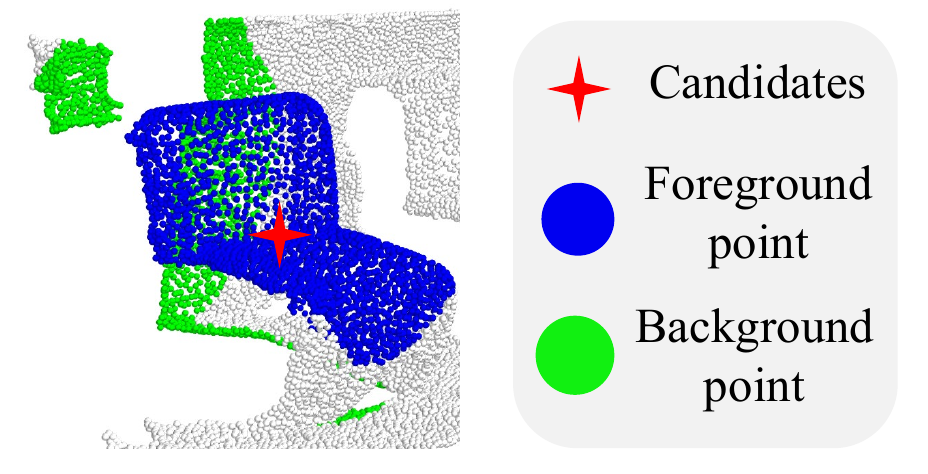}
   \caption{Shape modeling.}
   \label{fig:visualization-shape}
\end{figure}

To obtain a discriminative instance kernel, we encode semantic, positional and shape information to represent instance. The position and semantic information comes from candidate coordinates and point features. Shape information is thus encoded by splitting `foreground points' and `background points'. As illustrated in Fig.~\ref{fig:visualization-shape}, `foreground points' sketch a basic shape of instance (a chair). In Table~\ref{table:ablation-info-component} we show the performance when the position (coordinates) or shape (mixing foreground and background points for feature pooling) is ablated. One can observe that both information is vital for instance kernels.
\begin{table}
\begin{minipage}{0.48\linewidth}
\centering
\caption{Component analysis.}

\resizebox{1.0\linewidth}{!}{
\label{table:ablation-info-component}
\begin{tabular}{l|ccc}
\toprule
Info. components & $mAP$ & $AP@50$ & $AP@25 $\\
\midrule
W/o coord. & $49.5$ & $66.3$  & $75.9$ \\
W/o shape & $49.3$ & $64.3$  & $75.2$ \\
W/o both & $47.9$ & $63.6$  & $73.8$ \\
Full & 50.8 & 66.7 & 76.9\\
\bottomrule
\end{tabular}}
\end{minipage}
\begin{minipage}{0.48\linewidth}
\centering

\caption{Kernel shape analysis.} 

\resizebox{1.0\linewidth}{!}{
\label{table:ablation-kernel-shape}
\begin{tabular}{l|ccc}
\toprule
Kernel shape/size & mAP & AP@50 & AP@25 \\ \midrule
$[8, 1]/169$ & 50.6 & 67.8 & 76.4 \\ 
$[16, 1]/337$ & 50.8 & 66.7 & 76.9  \\
$[16, 8,1]/465$ & 51.2 & 67.1 & 77.0 \\
$[16, 16,1]/609$ & 50.9 & 66.0 & 76.5 \\\bottomrule
\end{tabular}}

\end{minipage}
\end{table}

\begin{figure}[!t]
\centering
\setlength{\abovecaptionskip}{0.cm}
\includegraphics[width=1\linewidth]{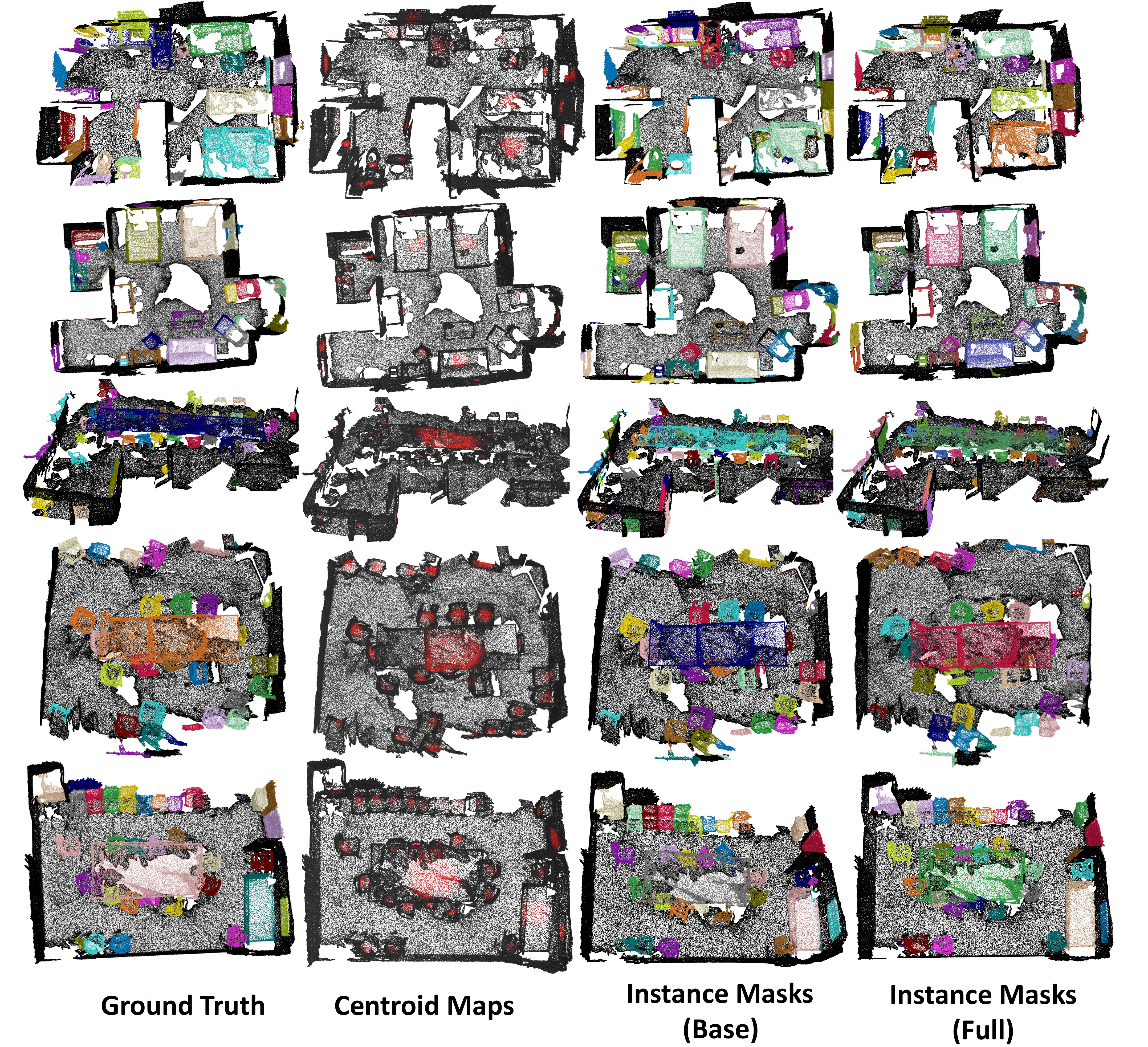}
\caption{\textbf{Visualizations on ScanNetV2 validation set.}}
\label{fig:visualization}
\end{figure}
\subsection{Dynamic Convolution}
To generate instance masks, point features are fed into different instance decoders consisting of a few dynamic convolution layers. The parameters of dynamic convolutions are conditioned on the corresponding instance kernels. As shown in Fig.~\ref{fig:convolution}, we instantiate the instance decoder with two convolution layers, which have $16$ and $1$ output channels (its kernel shape is $[16, 1]$). The elements in one instance kernel are sequentially inserted into the weight vectors and biases of these two convolution layers. Hence, the length of instance kernels $L$ depends on the specific configuration of the instance decoder. As for the instance decoder in Fig.~\ref{fig:convolution}, $L$ can be computed by:
\begin{align}
\label{eq: weight-length-detail}
    &Conv1\ \#weight = (16+3)\times 1\times 1\times 16=304,\ \#bias = 16\times 1=16\,,\\
    &Conv2\ \#weight = 16\times 1\times 1\times 1=16,\ \#bias = 1\times 1=1\,,\\
    &L = 304+16 + 16+1=337\,,
\end{align}
To evaluate the effect of kernel size, we design a series of ablation experiments. As illustrated in Table~\ref{table:ablation-kernel-shape}, DKNet is stable under varying instance kernel sizes and dynamic convolution layer shapes.

\subsection{Thresholds for Soft Instance Mask}
As mentioned in Section~3.5, in post-processing, we use the Otsu algorithm~\cite{otsu} to binarize the predicted soft instance masks. Otsu algorithm divides the pixels in a grey image into the foreground or the background category, whose essential idea is to maximize the inter-class variance. We re-purpose this idea to binarize the soft instance masks $M \in \mathbb{R}^{I\times N}$. As the original algorithm functions on pixels with discrete gray levels, similarly, we discreteize the value ($[0,1]$) of soft instance mask into $K$ confidence levels. Therefore, the quantified instance mask $M'$ can be obtained by:
\begin{equation}
    M'_{k} = \lfloor M_{k} * K \rfloor \,,
\end{equation}
where $M_k$ denotes the $k^{th}$ instance mask. Taking the quantified masks as inputs, Otsu algorithm processes each point in $M'_k$ as a pixel in an image, and outputs $T_{m,k}$ for each instance. Instead of using fixed threshold, Otsu algorithm can adaptively generate thresholds, which can better preserve the shape of instance with weak responses. 


\subsection{More Visualizations}
More visualizations of instance segmentation results and intermediate centroid maps are shown in Fig.~\ref{fig:visualization}. \textbf{Base} denotes the baseline method without candidate aggregation while \textbf{Full} denotes our full method. We also observe that, by reconstructing instance masks from instance kernels, some errors in semantic predictions can be corrected. We show some examples in Fig~\ref{fig:visualization-sem}.

\textbf{\begin{figure}[!t]
\centering
\setlength{\abovecaptionskip}{0.cm}
\includegraphics[width=1\linewidth]{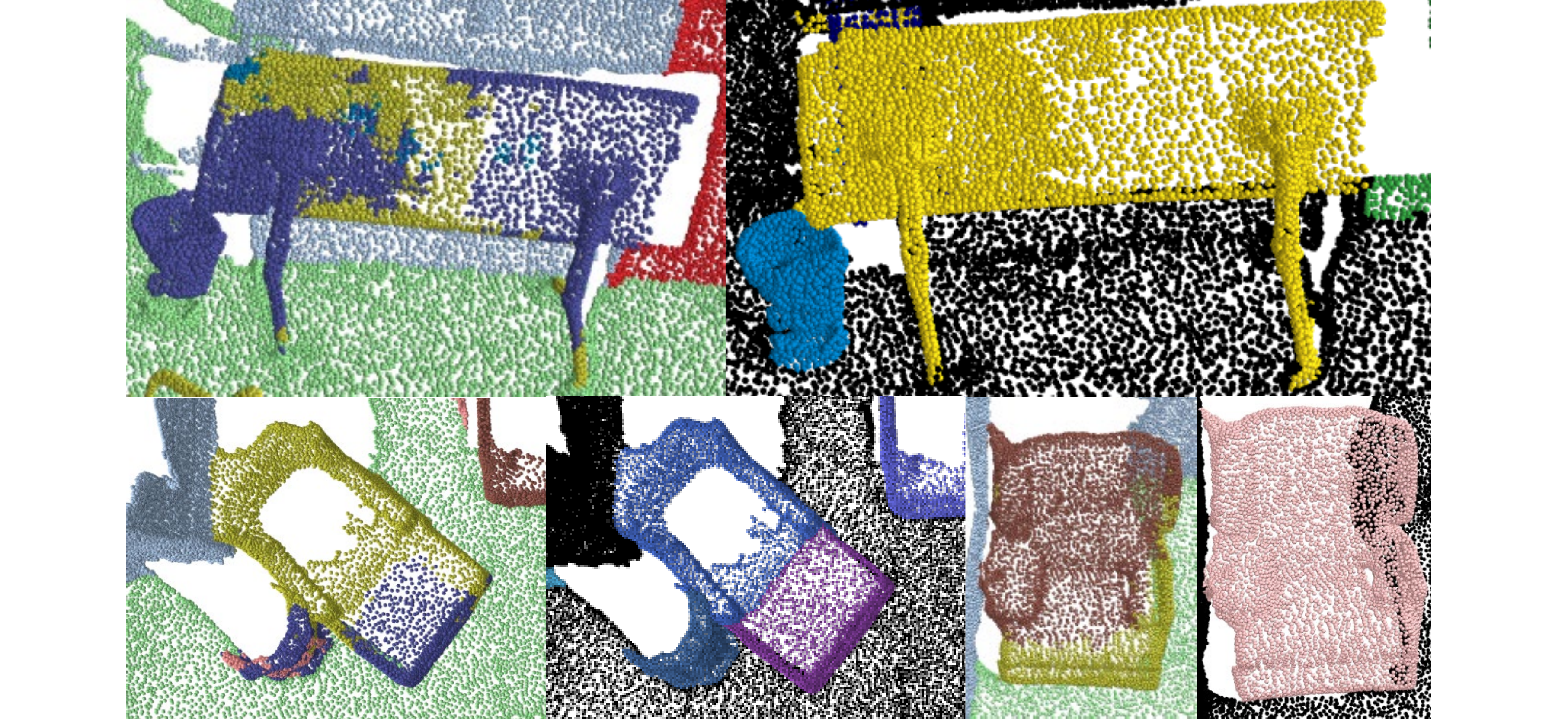}
\caption{\textbf{The robustness against errors in semantic predictions.} Although errors occur in the semantic predictions (left parts), the instance decoder can still recover correct instance masks (right parts).}
\label{fig:visualization-sem}
\end{figure}}

\subsection{Efficiency}
Here we specify the training and inference time of the proposed DKNet. Training DKNet on ScanNetV2~\cite{scannet} with default settings consumes about $72$ GPU hours on an single RTX 3090. In terms of inference, DKNet is relatively efficient; the average inference time (per scene) of DKNet on a Titan XP is $521$ ms, which is on par with recent bottom-up approaches PointGroup~\cite{pointgroup} ($452$ ms) and HAIS~\cite{hais} ($339$ ms) on the same device, only introducing limited latency ($100$-$200$ ms). Note that, DKNet is much more efficient than recent top-down approach such as GICN~\cite{gicn} ($8615$ ms). 

The detailed inference time of different steps in DKNet is shown Table~\ref{table:dknet-inference-time}. The core operation: encoding instance kernels consumes the second most time ($36.19\%$). To further boost the efficiency, we build a CUDA library for core operations in DKNet, \textit{e.g.}, the LN-NMS and Otsu thresholding mentioned in Sec.~3.4 and 3.5. Compared with naive python implementation, the CUDA version LN-NMS reduces its inference time from $187$ ms to $79$ ms. Note that, these operations are parallelized with naive implementations, with much room to be optimized. Hence, the inference time can probably be further reduced.

\begin{table}[t] 
\centering
\caption{Inference time of different stages in DKNet on RTX 3090.}  
\label{table:dknet-inference-time}
\begin{tabular}{ccccc}
\toprule
Total & Backbone & Encoding & Decoding & Post-processing \\ \midrule
357.5ms & 160.6ms & 129.3ms & 16.3ms & 45.9ms \\ 
\bottomrule
\end{tabular}
\end{table}  

\subsection{Full Evaluation Results}
The results under $AP@50$ and $mAP$ metrics on ScanNetV2 benchmark are reported in Table~\ref{table:performance_scan_ap50} and Table~\ref{table:performance_scan_map}.

\begin{table}[!t]
\scriptsize
\centering
\setlength{\tabcolsep}{1.5pt}
\renewcommand\arraystretch{1.2}
\caption{\textbf{Full quantitative results of $AP@50$ on the ScanNetV2 test set.} Best performance is in boldface.}
\label{table:performance_scan_ap50}

\resizebox{1\linewidth}{!}{
\begin{tabular}{@{}l|c|cccccccccccccccccc@{}}
\toprule
approaches&\rotatebox{90}{$AP@50$}&\rotatebox{90}{bathtub}&\rotatebox{90}{bed}&\rotatebox{90}{booksh.}&\rotatebox{90}{cabinet}&\rotatebox{90}{chair}&\rotatebox{90}{counter}&\rotatebox{90}{curtain}&\rotatebox{90}{desk}&\rotatebox{90}{door}&\rotatebox{90}{otherfu.}&\rotatebox{90}{picture}&\rotatebox{90}{refrige.}&\rotatebox{90}{s. curtain}&\rotatebox{90}{sink}&\rotatebox{90}{sofa}&\rotatebox{90}{table}&\rotatebox{90}{toilet}&\rotatebox{90}{window}\\
\midrule
3D-BoNet\cite{bonet}&48.8&\bf{100.0}&67.2&59.0&30.1&48.4&9.8&62.0&30.6&34.1&25.9&12.5&43.4&79.6&40.2&49.9&51.3&90.9&43.9\\
MTML\cite{mtml}&54.9&\bf{100.0}&80.7&58.8&32.7&64.7&0.4&81.5&18.0&41.8&36.4&18.2&44.5&\bf{100.0}&44.2&68.8&57.1&\bf{100.0}&39.6\\
3D-MPA\cite{3dmpa}&61.1&\bf{100.0}&83.3&76.5&52.6&75.6&13.6&58.8&47.0&43.8&43.2&35.8&65.0&85.7&42.9&76.5&55.7&\bf{100.0}&43.0\\
PointGroup\cite{pointgroup}&63.6&\bf{100.0}&76.5&62.4&50.5&79.7&11.6&69.6&38.4&44.1&55.9&47.6&59.6&\bf{100.0}&66.6&75.6&55.6&99.7&51.3\\
GICN\cite{gicn}&63.8&\bf{100.0}&\bf{89.5}&80.0&48.0&67.6&14.4&73.7&35.4&44.7&40.0&36.5&\bf{70.0}&\bf{100.0}&56.9&\bf{83.6}&59.9&\bf{100.0}&47.3\\
DyCo3D\cite{dyco3d}&64.1&\bf{100.0}&84.1&89.3&53.1&80.2&11.5&58.8&44.8&43.8&53.7&43.0&55.0&85.7&53.4&76.4&65.7&98.7&56.8\\
Occuseg\cite{occuseg}&67.2&\bf{100.0}&75.8&68.2&57.6&84.2&\bf{47.7}&50.4&52.4&56.7&58.5&45.1&55.7&\bf{100.0}&75.1&79.7&56.3&\bf{100.0}&46.7\\
SSTNet\cite{sstnet}&69.8&\bf{100.0}&69.7&\bf{88.8}&55.6&80.3&38.7&62.6&41.7&55.6&58.5&70.2&60.0&\bf{100.0}&\bf{82.4}&72.0&69.2&\bf{100.0}&50.9\\
HAIS\cite{hais}&69.9&\bf{100.0}&84.9&82.0&\bf{67.5}&80.8&27.9&\bf{75.7}&46.5&51.7&\bf{59.6}&55.9&60.0&\bf{100.0}&65.4&76.7&67.6&99.4&56.0\\
\bf{Ours}  &\bf{71.8}&\bf{100.0}&81.4&78.2&61.9&\bf{87.2}&22.4&75.1&\bf{56.9}&\bf{67.7}&58.5&\bf{72.4}&63.3&98.1&51.5&81.9&\bf{73.6}&\bf{100.0}&\bf{61.7}\\
\bottomrule
\end{tabular}}
\end{table}
\setlength{\tabcolsep}{1.4pt}
\begin{table}[!t] 
\scriptsize
\centering
\renewcommand\arraystretch{1.2}
\setlength{\tabcolsep}{1.5pt}
\caption{\textbf{Full quantitative results of $mAP$ on the ScanNetV2 test set.} Best performance is in boldface.}
\label{table:performance_scan_map}
\resizebox{1\linewidth}{!}{
\begin{tabular}{@{}l|c|cccccccccccccccccc@{}}
\toprule
approaches&\rotatebox{90}{$mAP$}&\rotatebox{90}{bathtub}&\rotatebox{90}{bed}&\rotatebox{90}{booksh.}&\rotatebox{90}{cabinet}&\rotatebox{90}{chair}&\rotatebox{90}{counter}&\rotatebox{90}{curtain}&\rotatebox{90}{desk}&\rotatebox{90}{door}&\rotatebox{90}{otherfu.}&\rotatebox{90}{picture}&\rotatebox{90}{refrige.}&\rotatebox{90}{s. curtain}&\rotatebox{90}{sink}&\rotatebox{90}{sofa}&\rotatebox{90}{table}&\rotatebox{90}{toilet}&\rotatebox{90}{window}\\
\midrule
3D-BoNet\cite{bonet}&25.3&51.9 &32.4 &25.1 &13.7 &34.5 &3.1 &41.9 &6.9 &16.2 &13.1 &5.2 &20.2 &33.8 &14.7 &30.1 &30.3 &65.1 &17.8\\
MTML\cite{mtml}&28.2&57.7 &38.0 &18.2 &10.7 &43.0 &0.1 &42.2 &5.7 &17.9 &16.2 &7.0 &22.9 &51.1 &16.1 &49.1 &31.3 &65.0 &16.2\\
3D-MPA\cite{3dmpa}&35.5&45.7 &48.4 &29.9 &27.7 &59.1 &4.7 &33.2 &21.2 &21.7 &27.8 &19.3 &41.3 &41.0 &19.5 &57.4 &35.2 &84.9 &21.3\\
PointGroup\cite{pointgroup}&40.7&63.9 &49.6 &41.5 &24.3 &64.5 &2.1 &\bf{57.0} &11.4 &21.1 &35.9 &21.7 &42.8 &66.0 &25.6 &56.2 &34.1 &86.0 &29.1\\
GICN\cite{gicn}&34.1&58.0 &37.1 &34.4 &19.8 &46.9 &5.2 &56.4 &9.3 &21.2 &21.2 &12.7 &34.7 &53.7 &20.6 &52.5 &32.9 &72.9 &24.1\\
DyCo3D\cite{dyco3d}&39.5&64.2 &51.8 &44.7 &25.9 &66.6 &5.0 &25.1 &16.6 &23.1 &36.2 &23.2 &33.1 &53.5 &22.9 &58.7 &43.8 &85.0 &31.7\\
Occuseg\cite{occuseg}&48.6&80.2 &53.6 &42.8 &36.9 &70.2 &\bf{20.5} &33.1 &30.1 &37.9 &47.4 &32.7 &43.7 &\bf{86.2} &48.5 &60.1&39.4 &84.6 &27.3\\
SSTNet\cite{sstnet}&50.6&73.8 &54.9 &49.7 &31.6 &69.3 &17.8 &37.7 &19.8 &33.0 &46.3 &57.6 &51.5 &85.7 &\bf{49.4} &63.7 &45.7 &\bf{94.3} &29.0\\
HAIS\cite{hais}&45.7&70.4 &56.1 &45.7 &36.4&67.3&4.6 &	54.7 &19.4 &30.8 &42.6 &28.8&45.4 &71.1 &26.2 &56.3 &43.4 &	88.9&34.4\\
\bf{Ours}  &\bf{53.2}&\bf{81.5} &\bf{62.4} &\bf{51.7} &	\bf{37.7} &	\bf{74.9} &	10.7 &	50.9 &	\bf{30.4} &	\bf{43.7} &	\bf{47.5} &	\bf{58.1} &	\bf{53.9} &	77.5 &	33.9 &	\bf{64.0} &	\bf{50.6} &90.1 &\bf{38.5}\\
\bottomrule
\end{tabular}}
\end{table}
\setlength{\tabcolsep}{1.4pt}

\end{document}